\documentclass[times,twocolumn,final]{elsarticle}

\usepackage{medima}
\usepackage{framed,multirow}
\usepackage{amssymb}
\usepackage{latexsym}
\usepackage{url}
\usepackage[dvipsnames]{xcolor}
\usepackage[colorlinks]{hyperref}
\usepackage{bbding}
\usepackage{soul}
\usepackage{bibunits}
\usepackage{bm}
\usepackage{natbib}
\usepackage{array}
\usepackage{tabularray}
\usepackage{graphicx}
\usepackage{subfigure}
\usepackage{pdflscape}
\usepackage{makecell}
\usepackage{adjustbox}
\usepackage{framed,multirow}
\usepackage{threeparttable}
\usepackage{amssymb}
\usepackage{pifont}
\usepackage{algorithm}
\usepackage{layouts}
\usepackage{listings}
\usepackage{colortbl}
\usepackage{booktabs}
\usepackage{amsmath}
\usepackage{pythonhighlight}
\usepackage{fancyhdr}
\usepackage{pifont}
\makeatletter
\newcommand\footnoteref[1]{\protected@xdef\@thefnmark{\ref{#1}}\@footnotemark}
\makeatother

\newcolumntype{P}[1]{>{\centering\arraybackslash}p{#1}}
\newlength\savewidth

\def\arrvline{\hfil\kern\arraycolsep\vline\kern-\arraycolsep\hfilneg}
\definecolor{Highlight}{HTML}{39b54a}

\definecolor{iblue}{rgb}{0.06, 0.75, 1.0}
\definecolor{igray}{rgb}{0.00, 0.00, 0.00}

\definecolor{significant}{RGB}{239,134,131}

\definecolor{newcolor}{rgb}{.8,.349,.1}

\journal{Knowledge-Based Systems}

\begin{document}
	
	\verso{ \textit{et~al.}}
	
	\begin{frontmatter}
 
		\title{SP-Det: Self-Prompted Dual-Text Fusion for Generalized Multi-Label Lesion Detection}
		
		\author[1,2]{Qing~Xu$^{\dag,}$}
		\author[1]{Yanqian~Wang$^{\dag,}$}
		\author[1]{Xiangjian~He\corref{cor1}}
        \ead{sean.he@nottingham.edu.cn}
		\author[1]{Yue~Li}
        \author[1]{Yixuan~Zhang}
		\author[2]{Rong~Qu}
        \author[3]{Wenting~Duan}
		\author[4]{Zhen~Chen}
		
		\cortext[cor1]{Corresponding author. \dag{Equal contribution.}}
		
		\address[1]{School of Computer Science, University of Nottingham Ningbo China, Ningbo, 31500, Zhejiang, China}
        \address[2]{School of Computer Science, University of Nottingham, University Park, Nottingham, NG7 2RD, UK}
		\address[3]{School of Computer Science, University of Lincoln, Lincoln  LN6 7TS, UK}
		\address[4]{Yale University, New Haven, CT 06510, USA.}

		\received{-}
		\finalform{-}
		\accepted{-}
		\availableonline{-}
		
\begin{abstract}
Automated lesion detection in chest X-rays has demonstrated significant potential for improving clinical diagnosis by precisely localizing pathological abnormalities. While recent promptable detection frameworks have achieved remarkable accuracy in target localization, existing methods typically rely on manual annotations as prompts, which are labor-intensive and impractical for clinical applications. To address this limitation, we propose SP-Det, a novel self-prompted detection framework that automatically generates rich textual context to guide multi-label lesion detection without requiring expert annotations. Specifically, we introduce an expert-free dual-text prompt generator (DTPG) that leverages two complementary textual modalities: semantic context prompts that capture global pathological patterns and disease beacon prompts that focus on disease-specific manifestations. Moreover, we devise a bidirectional feature enhancer (BFE) that synergistically integrates comprehensive diagnostic context with disease-specific embeddings to significantly improve feature representation and detection accuracy. Extensive experiments on two chest X-ray datasets with diverse thoracic disease categories demonstrate that our SP-Det framework outperforms state-of-the-art detection methods while completely eliminating the dependency on expert-annotated prompts compared to existing promptable architectures. 
\end{abstract}
		\begin{keyword}
    \KWD 
    Chest X-ray analysis \sep Lesion detection \sep Self-prompting \sep Multi-modal fusion
\end{keyword}
	\end{frontmatter}
 
\section{Introduction}\label{sec:introduction}

Medical imaging plays a vital role in disease prevention, diagnosis, and management across diverse clinical scenarios. Among various imaging modalities, chest X-rays (CXRs) represent the most widely utilized examination technique for thoracic disease assessment worldwide, including critical conditions such as COVID-19 \cite{toussie2020clinical}, owing to their cost-effectiveness, minimal radiation exposure, and rapid acquisition protocols \cite{cid2024development}. Despite their clinical prevalence, CXR interpretation traditionally demands the expertise of specialized radiologists, creating a resource-intensive process subject to inter-observer variability and workflow bottlenecks. With the exponentially increasing volume of CXR examinations globally \cite{rimmer2017radiologist,cid2024development}, there is an urgent need for automated lesion localization methods to enhance diagnostic efficiency, reduce interpretation time, and improve consistency in clinical decision-making.

\begin{figure*}[!t]
    \centering
    \includegraphics[width=1\linewidth]{images/fig_intro.png}
    \caption{Comparison of our self-prompted detection with existing lesion detection paradigms in chest X-rays. (1) Non-prompted detection: Traditional models using only image-label pairs without expert knowledge. (2) Expert-prompted detection: Multimodal models requiring manually curated disease categories or clinical reports. (3) Self-prompted detection (SP-Det): Our approach automatically generates semantic context and disease beacons from CXR images using a medical vision-language model (VLM), eliminating manual expert annotation requirements.}
    \label{fig:detection_paradigms_1}
\end{figure*}

In this challenging context, classical lesion detection methods are typically categorized into two primary paradigms: two-stage detectors and one-stage detectors \cite{zou2023object}. Two-stage detectors prioritize accuracy through region proposal networks that first generate potential object locations before performing classification and refinement, achieving superior detection precision at the cost of computational redundancy due to their sequential processing pipeline. In contrast, one-stage detectors, exemplified by the You Only Look Once (YOLO) series \cite{redmon2016you}, identify all targets in a single forward pass by directly predicting from feature maps, offering faster inference speeds suitable for real-time applications. However, as illustrated in Fig.~\ref{fig:detection_paradigms_1}(a), such frameworks rely exclusively on visual input while overlooking the complementary semantic information embedded in textual descriptions, which limits their ability to leverage rich diagnostic context available in clinical settings.

Recognizing the potential of cross-modal information fusion, multimodal detection models \cite{zareian2021open,gu2021open,cheng2024yolo} have emerged as a promising paradigm that aligns vocabulary embeddings with image features through contrastive learning, as shown in Fig.~\ref{fig:detection_paradigms_1}(b). This detection paradigm substantially improves the generalization capabilities of models by expanding the label space beyond predefined categories, enabling zero-shot or few-shot detection of novel object classes through semantic relationships encoded in language models. Although demonstrably beneficial, existing multimodal detection approaches typically rely on expert-curated predefined categorical vocabularies or manually annotated grounding texts during training, which perpetuates a significant bottleneck in terms of scalability and practicality. This reliance on labor-intensive annotation processes substantially limits the applicability to real-world clinical scenarios, where textual context should ideally be dynamically derived from medical imaging data without expert intervention.

To overcome this limitation, we propose SP-Det, a novel self-prompting detection framework that automatically generates and leverages rich textual context to guide multi-label lesion detection, as illustrated in Fig.~\ref{fig:detection_paradigms_1}(c). Our approach eliminates the dependency on expert annotations by implementing an expert-free dual-text prompt generator (DTPG) that automatically derives complementary textual guidance directly from imaging data. Specifically, the DTPG leverages a pre-trained Vision-Language Model (VLM) to synthesize two distinct but synergistic textual modalities: semantic context prompts (SCP) that provide holistic understanding of the entire radiograph, and disease beacon prompts (DBP) that offer targeted guidance on specific pathological findings. These complementary textual representations capture different aspects of diagnostic information, where SCP establishes global contextual awareness while DBP focuses on disease-specific characteristics. Furthermore, we introduce a bidirectional feature enhancer (BFE) that intelligently integrates these complementary textual representations with visual features through cross-modal attention mechanisms. The synergistic utilization of global diagnostic context alongside disease-specific embeddings significantly enhances feature representation capabilities, leading to more precise lesion localization and robust classification performance. Extensive experiments on the VinDr-CXR dataset demonstrate substantial and consistent performance improvements over existing state-of-the-art methods in multi-label lesion detection of CXRs, validating the effectiveness of our self-prompting paradigm. To our knowledge, SP-Det represents the first framework that fully automates the prompt generation pipeline for lesion detection in CXRs, leveraging pre-trained medical VLMs to synthesize contextual and category-specific guidance derived from visual content alone without requiring any manual annotations. This addresses a critical gap in making multimodal detection more practical and scalable for real-world clinical deployment, where expert annotation resources are often limited or unavailable.

Our contributions can be summarized as follows:
\begin{itemize}
    \item We propose SP-Det, a novel self-prompting framework for multi-label lesion localization in CXRs that eliminates the dependency on expert annotations, establishing a new paradigm for automated prompt generation in medical image analysis. To the best of our knowledge, this is the first work to fully automate the prompt generation pipeline for lesion detection using pre-trained medical VLMs.
    \item We devise the DTPG that automatically synthesizes two complementary textual modalities: SCP for holistic radiograph understanding and DBP for targeted pathological guidance, enabling rich cross-modal supervision without manual intervention.
    \item We introduce the BFE that significantly improves feature representation capabilities through synergistic integration of comprehensive diagnostic context and disease-specific embeddings via cross-modal attention mechanisms, seamlessly incorporating into the detection pipeline to achieve robust cross-modal feature fusion.
    \item Extensive experiments on the two datasets with diverse disease categories demonstrate that our SP-Det framework achieves consistent and substantial performance improvements over state-of-the-art detection methods, validating its effectiveness and practical value for scalable, expert-free multi-label lesion detection in clinical scenarios.
\end{itemize}

\section{Related Work}
\label{sec:related_work}

\subsection{Chest X-ray Analysis}
Deep learning methods have significantly advanced the automatic analysis of chest X-ray images, evolving from basic classification to sophisticated localization tasks. Initial research primarily focused on image-level classification for disease recognition. Tang et al. \cite{tang2020automated} conducted a comprehensive comparative analysis of various convolutional neural network (CNN) architectures for binary classification, distinguishing between normal and abnormal CXR images. Subsequently, numerous approaches have been developed for recognizing specific disease patterns, including tuberculosis \cite{zhao2021diagnose} and pneumonia \cite{ismael2021deep}. However, these classification-based methods face fundamental limitations in clinical practice, as multiple pathological abnormalities frequently coexist in chest X-ray examinations, necessitating multi-label recognition capabilities. This clinical requirement has driven recent research toward multi-label classification frameworks. For instance, CNN-ELM \cite{nahiduzzaman2023parallel} introduced a lightweight model architecture capable of classifying 17 distinct lesion categories simultaneously. CheXFusion \cite{kim2023chexfusion} leveraged Transformer-based attention mechanisms to fuse multi-view CXR images, effectively addressing the long-tail distribution problem inherent in multi-label classification tasks.

While image-level classification successfully indicates the presence of certain diseases, it fundamentally fails to localize specific lesions, which is critical for clinical interpretability and diagnostic confidence. Object detection methods overcome this limitation by simultaneously predicting both the presence and precise spatial location of pathological abnormalities. Fan et al. \cite{fan2024deep} proposed a YOLOX-based detection framework capable of identifying 14 abnormality categories, accompanied by a large-scale annotated benchmark dataset CXR-AL14. Additionally, various detection architectures with different backbone networks have been systematically explored to optimize performance in chest X-ray analysis \cite{van2024exploring}. Despite these advances, existing detection methods predominantly rely on visual features alone, overlooking the potential benefits of incorporating rich semantic information from textual modalities. Our SP-Det framework addresses this gap by automatically generating and integrating textual context to guide lesion detection, establishing a more comprehensive multimodal detection paradigm.

\subsection{Vision-Language Models for Object Detection}
Vision-language models (VLMs) have demonstrated remarkable zero-shot generalization capabilities across diverse visual recognition tasks through cross-modal learning \cite{jia2021scaling}. Contrastive learning approaches, exemplified by CLIP \cite{radford2021learning}, learn discriminative joint representations by optimizing the alignment between paired image-text samples in a shared embedding space while maximizing the separation of unpaired samples. Region-text alignment methods establish correspondences between specific image regions and corresponding textual phrases \cite{li2022grounded}, enabling fine-grained cross-modal understanding. Some pioneering works distill knowledge from large-scale pre-trained VLMs into task-specific detection models \cite{gu2021open, zareian2021open}, although their effectiveness remains constrained by the predefined vocabulary of the teacher models. To overcome vocabulary limitations and expand the detection label space beyond predefined categories, open-vocabulary object detection (OVOD) has emerged as a promising research direction. OVR-CNN \cite{zareian2021open} employs large-scale image-caption pairs to train a ResNet-based backbone network integrated with a vision-to-language mapping module, enabling semantic alignment between visual and textual representations. The GLIP family \cite{li2022grounded, zhang2022glipv2} pioneered the unification of phrase grounding and object detection through a shared pre-training framework, achieving substantial improvements in cross-modal understanding. Grounding DINO \cite{liu2024grounding} further advanced this paradigm by combining grounding pre-training with detection transformers, significantly enhancing cross-modal alignment capabilities and detection performance.

In the medical imaging domain, automated report generation models serve as specialized VLMs that produce diagnostic interpretations directly from radiological images \cite{reale2024vision}. While early approaches predominantly adopted encoder-decoder architectures, recent methodologies leverage large language models (LLMs) as powerful medical text generators. Notable examples include CheXagent \cite{chen2024chexagent}, which demonstrated a 36\% reduction in reporting time for radiology residents. However, existing vision-language detection frameworks typically require expert-curated vocabularies or manually annotated grounding texts, limiting their scalability and practical applicability in real-world clinical scenarios. In contrast, our SP-Det framework leverages pre-trained medical VLMs to automatically generate rich textual prompts without expert annotations, effectively guiding multi-label lesion detection while eliminating the dependency on labor-intensive manual supervision.

\section{Methodology}
\label{sec:method}

\begin{figure*}[!t]
    \centering
    \includegraphics[width=0.95\linewidth]{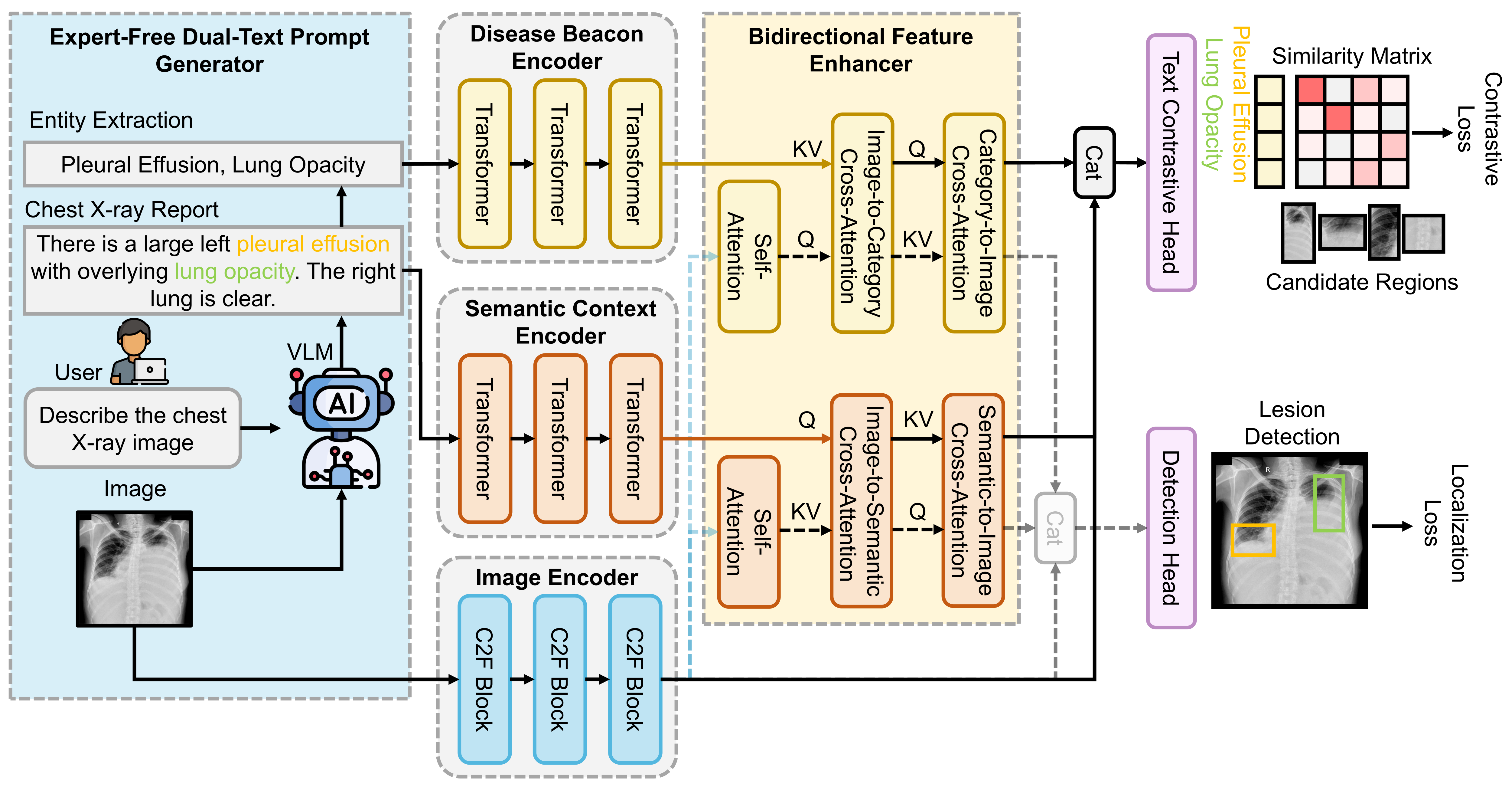}
    \caption{Overview of the SP-Det framework for multi-label lesion detection. The framework comprises four two components: (1) Expert-free dual-text prompt generator: A medical vision-language model automatically generates clinically relevant reports from CXR images to provide semantic context prompts, while disease categories extracted from these reports serve as disease beacon prompts. (2) Bidirectional feature enhancer: The process begins with self-attention enhancement of the highest-level image features $X_h$, followed by bidirectional cross-attention with text features where each modality alternately functions as queries (Q) and keys/values (K, V). Finally, these refined high-level features are channel-wise concatenated with the original low-level image features $X_l$, ensuring both semantic understanding and spatial detail preservation.}
    \label{fig:architecture}
\end{figure*}

The architecture of our proposed SP-Det framework is illustrated in Fig. \ref{fig:architecture}. Given chest X-ray images, SP-Det automatically generates complementary textual guidance for multi-label lesion detection, eliminating the dependency on manual expert annotations. The framework incorporates two key innovations: (1) DTPG that autonomously produces clinically relevant SCP and DBP directly from input CXRs, and (2) BFE that enriches visual representations through cross-modal attention mechanisms guided by both semantic information and targeted disease signals. This integrated architecture enables synergistic utilization of dual textual modalities to enhance detection performance while maintaining full automation in the prompt generation process. In the following subsections, we provide detailed descriptions of each component.

\subsection{Expert-Free Dual-Text Prompt Generator}
\label{subsec:dtpg}

To eliminate the dependency on expert annotations while preserving rich diagnostic context, we devise an expert-free dual-text prompt generator that automatically synthesizes two complementary textual modalities from input chest X-ray images. This generator leverages pre-trained medical vision-language models to produce semantic context prompts that capture holistic radiological understanding and disease beacon prompts that provide targeted category-specific guidance. The automatic generation process ensures scalability and practical applicability in real-world clinical deployment scenarios.

\subsubsection{Semantic Context Prompt Generation}
To automatically generate high-quality semantic context prompts, we employ a Visual Question Answering (VQA) pipeline specifically tailored for chest X-ray interpretation. The pipeline utilizes a transformer-based image encoder from the SigLIP-Large architecture \cite{zhai2023sigmoid} for visual feature extraction, coupled with the Phi-2 decoder-only transformer \cite{li2023textbooks} for text generation. Phi-2, pre-trained on extensive clinical corpora including medical guidelines and research articles, possesses comprehensive medical and clinical knowledge essential for accurate radiological report generation. We adopt a single-round dialogue approach with the standardized prompt \textit{"Describe the chest X-ray image"} to ensure consistent report generation across all inputs, avoiding potential biases introduced by variable prompting strategies. The generated reports undergo systematic post-processing to enhance quality and coherence: incomplete sentences at paragraph endings are removed, and redundant text segments are filtered out to improve linguistic clarity. This automated process produces diagnostic reports that capture critical radiological findings, including disease presence, anatomical locations, and pathological characteristics, thereby providing rich semantic context for subsequent visual feature refinement without requiring manual annotations.

\subsubsection{Disease Beacon Prompt Extraction}
To eliminate labor-intensive expert labeling while maintaining disease-specific guidance, we adopt a language-based strategy to extract disease beacon prompts directly from the automatically generated diagnostic reports. Specifically, we employ a noun phrase extractor based on a pre-trained natural language processing (NLP) parser to identify candidate disease entities. Given a report sentence, all noun phrases are systematically extracted and treated as potential category labels for disease classification. To handle the inherent ambiguity in medical terminology, we implement semantic similarity matching using clinical word embeddings, allowing the model to recognize that terms such as \textit{opacity} and \textit{consolidation} may refer to similar or overlapping pathological findings. For each annotated bounding box in the training data, we compare its associated class label with the extracted noun phrases using semantic similarity scoring. If a noun phrase semantically matches the box label above a predefined threshold, we treat it as a positive region-text pair, establishing correspondence between visual regions and textual descriptions.

Importantly, noun phrases that do not correspond to actual lesions are intentionally retained as negative prompts, serving a crucial role in the training process by enabling the model to learn discriminative representations. Any spurious bounding boxes generated by redundant terms are effectively eliminated through Non-Maximum Suppression (NMS) post-processing during inference. Additionally, negation detection is incorporated to identify negative statements in reports (e.g., \textit{"no signs of pneumonia"}), ensuring that negated conditions are excluded from positive region-text pairs. This comprehensive approach enables our model to learn robust region-text alignment while handling the inherent complexities and ambiguities of medical language, without relying on expert supervision.

\subsubsection{Prompt Encoding Strategy}
SP-Det employs two specialized text encoders for prompt encoding, reflecting the distinct nature of the two textual modalities. This design choice is motivated by the fundamental differences in information representation: radiology reports provide descriptive information at the sentence level, capturing comprehensive diagnostic context, while disease names are concise categorical labels requiring precise alignment with visual features for classification. For semantic context prompts, we utilize a mono-modal text encoder to produce token-level embeddings that preserve the detailed linguistic structure and semantic relationships within diagnostic reports. For disease beacon prompts, we adopt a vision-language model encoder that aligns text and image representations in a shared embedding space, facilitating semantic correspondence between disease labels and visual features for contrastive learning. While these encoders operate in different dimensional spaces, the bidirectional cross-attention mechanism in our feature enhancer naturally addresses any scale unification concerns. Each text modality performs independent cross-attention with image features, where the attention operations produce outputs that maintain the original image feature dimensions through learnable projection matrices. This design ensures model stability by preserving consistent feature scales throughout the fusion process, eliminating the need for explicit embedding fusion between the two text encoders and maintaining architectural simplicity.

\subsection{Bidirectional Feature Enhancer}
\label{subsec:bfe}

To enable effective visual-textual fusion while preserving spatial information, we devise the bidirectional feature enhancer as the core fusion module of SP-Det. This enhancer operates through three sequential stages: (1) self-attention enhancement of visual features, (2) bidirectional cross-attention for multi-modal fusion, and (3) feature integration with dimensional preservation. Each stage is carefully designed to progressively refine visual representations by cross-modal interactions.

\subsubsection{Self-Attention Enhancement}
Our image encoder adopts the RepVL-PAN architecture \cite{cheng2024yolo}, producing multi-scale feature maps with progressively deeper semantic abstractions. Following findings from Grounding DINO \cite{liu2024grounding} that higher-level features enhance predictive robustness and semantic understanding, we apply the bidirectional enhancer exclusively to the highest-level feature map $X_h \in \mathbb{R}^{H \times W \times D_h}$, where $H$ and $W$ denote spatial dimensions and $D_h$ represents the feature dimension. Lower-level features $X_l$ are preserved for subsequent concatenation to maintain spatial precision. To prepare for cross-modal interaction, we flatten the spatial dimensions and incorporate positional encoding to preserve spatial relationships:
\begin{equation}
X_h^{flat} = \text{Flatten}(X_h) \in \mathbb{R}^{L \times D_h}, \quad L = H \times W,
\end{equation}
\begin{equation}
X_h^{pos} = X_h^{flat} + \text{PE}(X_h^{flat}),
\end{equation}
where $\text{PE}(\cdot)$ denotes learnable positional encoding. We then apply multi-head self-attention to adaptively emphasize relevant visual regions:
\begin{equation}
X_h^{self} = \mathcal{M}(Q=X_h^{pos}, K=X_h^{pos}, V=X_h^{flat}) + X_h^{flat},
\end{equation}
\begin{equation}
X_h^{refined} = \text{FFN}(\text{LayerNorm}(X_h^{self})) + X_h^{self},
\end{equation}
where $\mathcal{M}$ denotes multi-head attention, $\text{LayerNorm}$ represents layer normalization, and $\text{FFN}$ denotes a feed-forward network with residual connections. This self-attention mechanism enables the model to capture long-range dependencies within visual features before cross-modal fusion.
\begin{table*}[!t]
\centering
\caption{Comparison with state-of-the-art models on multi-label lesion detection.}
\setlength\tabcolsep{17pt}
\resizebox{1\textwidth}{!}{%
\begin{tabular}{c| c c c c c c c c c}
    \hline
    Methods & Precision & Recall & \textbf{$\rm AP_{\rm 40}$} & \textbf{$\rm AP_{\rm 50}$} & \textbf{$\rm AP_{\rm 60}$} & \textbf{$\rm AP_{\rm 70}$} & \textbf{$\rm AP_{\rm 80}$} & \textbf{$\rm AP_{\rm 90}$} & \textbf{$\rm mAP_{\rm 40:95}$} \\
    \hline
    RT-DETR \cite{zhao2024detrs} & 44.9 & 40.5 & 39.6 & 33.9 & 25.4 & 15.3 & 5.0 & 0.3 & 18.4 \\
    DualAttNet  \cite{xu2024dualattnet} & 45.5 & 40.2 & 39.1 & 34.8 & 26.5 & 16.8 & 6.5 & 0.5 & 19.1 \\
    MLLA \cite{yuan2024multi} & 45.3 & 40.4 & 38.9 & 33.3 & 25.6 & 16.2 & 5.1 & 0.4 & 18.5 \\
    YOLOv8 \cite{yolov8_ultralytics} & 45.7 & 41.2 & \underline{40.3} & 35.3 & 28.0 & 15.4 & \underline{6.6} & 0.8 & 19.4 \\
    YOLOv9 \cite{wang2024yolov9} & 43.6 & \underline{42.5} & 38.1 & 33.5 & 26.0 & 16.9 & 6.3 & 0.7 & 18.7 \\
    YOLOv10 \cite{wang2024yolov10} & 44.8 & 40.2 & 37.0 & 33.0 & 27.2 & 16.5 & 5.9 & \underline{1.7} & 18.7 \\
    YOLOv11 \cite{yolo11_ultralytics} & 45.0 & 41.0 & 39.4 & 34.7 & 27.3 & 15.8 & 5.6 & 0.7 & 18.9 \\
    YOLO-World \cite{cheng2024yolo} & \textbf{46.7} & 39.2 & 40.1 & \underline{35.6} & \underline{29.0} & \underline{18.3} & \textbf{7.4} & \textbf{2.5} & \underline{20.5} \\
    SP-Det (Ours) & \underline{46.4} & \textbf{43.8} & \textbf{42.4} & \textbf{38.2} & \textbf{29.5} & \textbf{18.7} & \underline{6.6} & 0.7 & \textbf{21.0} \\
    \hline
\end{tabular}%
}
\label{tab:model_comparison}
\end{table*}

\begin{table*}[!t]
\caption{Lesion-level comparison with state-of-the-art models ($\rm AP_{\rm 40}$\%).}
\tabcolsep=1em
\resizebox{\textwidth}{!}{
\begin{tabular}{c|ccccccccc}
    \hline
    Category & RT-DETR & DualAttNet & MLLA & YOLOv8 & YOLOv9 & YOLOv10 & YOLOv11 & YOLO-World & SP-Det \\
    \hline
    Aortic enlargement & 43.0 & 46.0 & 45.9 & 44.0 & 45.5 & 39.5 & 46.0 & \textbf{50.0} & \underline{49.6} \\
    Atelectasis & 48.0 & \underline{48.4} & 46.5 & \textbf{49.6} & 43.5 & 37.3 & 46.9 & 47.3 & 48.0 \\
    Calcification & 41.4 & 40.0 & 43.0 & 42.5 & 35.6 & 42.3 & 35.5 & \underline{45.2} & \textbf{54.3} \\
    Cardiomegaly & 69.2 & 70.0 & 68.1 & \textbf{73.1} & 71.0 & 69.4 & 71.4 & \underline{72.5} & 69.8 \\
    Consolidation & 32.1 & 31.2 & \underline{32.3} & 24.1 & 26.6 & 23.1 & 16.5 & 23.2 & \textbf{33.8} \\
    ILD & \underline{32.5} & 32.4 & 30.8 & 29.4 & 25.6 & 28.3 & 18.3 & 26.8 & \textbf{33.3} \\
    Infiltration & 33.6 & 32.8 & 36.0 & 33.8 & 37.6 & 33.0 & \textbf{41.5} & \underline{41.0} & 36.1 \\
    Lung Opacity & 20.6 & 23.7 & 23.2 & 23.4 & \underline{23.8} & 19.5 & 18.2 & 22.4 & \textbf{25.0} \\
    Nodule/Mass & 33.4 & 33.8 & 33.7 & 34.5 & \underline{35.6} & 30.9 & 31.8 & \textbf{36.3} & 34.8 \\
    Other lesion & 14.6 & 14.5 & 12.9 & 13.4 & 13.8 & 12.0 & 12.4 & \underline{14.6} & \textbf{15.6} \\
    Pleural effusion & 39.3 & 40.1 & 40.3 & \underline{45.9} & 44.3 & 45.1 & 43.2 & \textbf{51.0} & 42.4 \\
    Pleural thickening & 26.5 & 26.0 & 24.9 & 27.1 & 25.2 & 23.2 & 25.8 & \underline{27.5} & \textbf{28.5} \\
    Pneumothorax & 76.7 & \underline{79.3} & 73.3 & 73.9 & 46.5 & 53.9 & 79.0 & 63.5 & \textbf{86.5} \\
    Pulmonary fibrosis & 30.0 & 29.6 & 26.9 & \textbf{35.2} & 33.0 & 31.5 & \underline{34.0} & 31.5 & 32.6 \\
    \hline
\end{tabular}
}
\label{tab:disease_comparison}
\end{table*}

\subsubsection{Bidirectional Cross-Attention Fusion}
The bidirectional cross-attention mechanism enables mutual guidance between visual and textual modalities, allowing each modality to attend to relevant information from the other. Since both semantic context prompts and disease beacon prompts undergo identical fusion processes with image features, we use $T \in \mathbb{R}^{N_t \times D_t}$ to represent text embeddings generically, where $N_t$ denotes sequence length and $D_t$ represents the text embedding dimension. To enable cross-modal interaction, we first project both modalities to a shared embedding space through learnable transformation matrices:
\begin{equation}
\hat{X} = X_h^{refined} W_v, \quad \hat{T} = T W_t,
\end{equation}
where $W_v \in \mathbb{R}^{D_h \times D_{shared}}$ and $W_t \in \mathbb{R}^{D_t \times D_{shared}}$ are learnable projection matrices mapping to the unified dimension $D_{shared}$. The bidirectional fusion proceeds through two sequential cross-attention operations:

Image-to-Text Attention: Visual features query textual information to extract visually-relevant semantic context:
\begin{equation}
T_{guided} = \mathcal{M}(Q=\hat{X}, K=\hat{T}, V=\hat{T}),
\end{equation}

Text-to-Image Attention: The refined text features $T_{guided}$ guide visual attention toward diagnostically critical regions:
\begin{equation}
X_{cross} = \mathcal{M}(Q=T_{guided}, K=\hat{X}, V=\hat{X}).
\end{equation}
This bidirectional attention design ensures comprehensive multi-modal understanding by enabling visual features to attend to relevant textual context while simultaneously allowing text-guided attention to highlight critical visual regions for lesion localization.

\subsubsection{Feature Integration with Dimensional Preservation}
To maintain compatibility with the downstream detection pipeline, we project the cross-attended features back to the original visual feature dimension:
\begin{equation}
X_{proj} = X_{cross} W_p, \quad W_p \in \mathbb{R}^{D_{shared} \times D_h}.
\end{equation}
Then, a learnable scaling parameter $\gamma$ controls the contribution of cross-modal information through a residual connection, enabling adaptive fusion:
\begin{equation}
X_{enhanced} = \text{FFN}(\text{LayerNorm}(\gamma \cdot X_{proj} + X_h^{refined})).
\end{equation}
Finally, we restore the spatial structure and concatenate with preserved low-level features to maintain both semantic understanding and spatial precision:
\begin{equation}
X_{spatial} = \text{Reshape}(X_{enhanced}) \in \mathbb{R}^{H \times W \times D_h},
\end{equation}
\begin{equation}
X_{final} = \text{Concat}(X_l, X_{spatial}).
\end{equation}

This architectural design strategically maintains spatial precision through preserved low-level features while incorporating rich semantic guidance from high-level cross-modal fusion, enabling both accurate localization and comprehensive contextual understanding for multi-label lesion detection. The bidirectional attention mechanism seamlessly handles text features of varying dimensions from two distinct text encoders without affecting the output image feature dimensionality, ensuring stable integration with the downstream optimization pipeline.

\begin{figure*}[!t]
    \centering
    \includegraphics[width=1\linewidth]{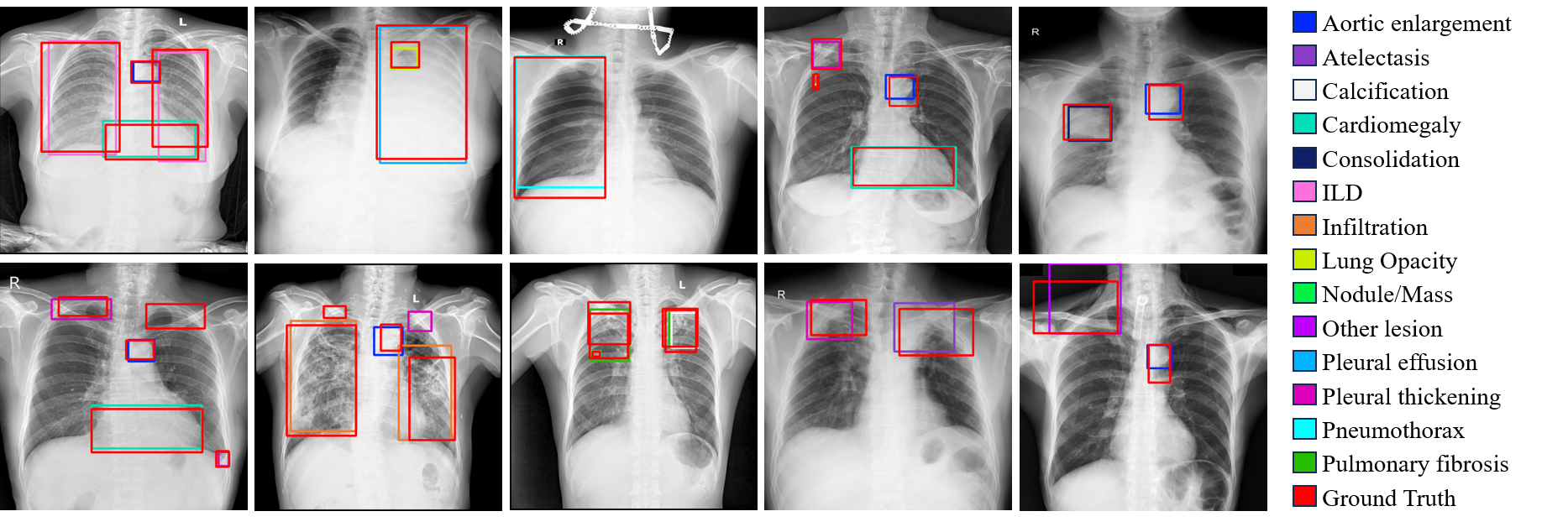}
    \caption{Case study from the VinDr-CXR test set. The predicted bounding boxes are compared with the ground truth annotations, with each disease highlighted using a distinct colour, while the ground truth is marked with a red box.}
    \label{fig:visualisation}
\end{figure*}

\begin{table*}[!t]
    \centering
    \small
    \setlength\tabcolsep{11pt}
    \caption{Zero-shot lesion-level comparison on the ChestX-ray8 dataset ($\rm AP_{40}$\%).}
    \resizebox{\textwidth}{!}{
    \begin{tabular}{c|ccccccccc}
    \hline
     Disease & RT-DETR & DualAttNet & MLLA & YOLOv8 & YOLOv9 & YOLOv10 & YOLOv11 & YOLO-World & SP-Det \\
    \hline
    Atelectasis & 8.2 & 7.6 & 9.1 & 8.8 & 9.4 & 7.9 & 8.6 & \underline{10.3} & \textbf{12.8} \\
    Cardiomegaly & 52.3 & 54.1 & 48.2 & 58.7 & 56.8 & 59.2 & \underline{60.1} & 58.4 & \textbf{63.9} \\
    Effusion & 24.8 & 26.2 & 25.9 & 28.1 & 27.6 & 26.4 & 27.2 & \underline{31.5} & \textbf{35.2} \\
    Infiltration & 19.5 & 18.8 & 21.2 & 20.1 & 22.4 & 18.9 & \textbf{24.6} & \underline{23.8} & 23.2 \\
    Mass & 16.8 & 18.2 & 17.4 & 19.5 & \textbf{20.1} & 17.3 & 18.6 & \underline{19.8} & 19.5 \\
    Nodule & 18.3 & 19.1 & 18.8 & 20.2 & 21.5 & 17.9 & 19.4 & \underline{22.8} & \textbf{26.1} \\
    Pneumonia & 21.7 & 20.4 & 23.8 & 22.3 & 25.1 & 21.5 & \underline{27.2} & 26.4 & \textbf{29.8} \\
    Pneumothorax & 42.6 & 45.3 & 41.1 & 44.8 & 35.2 & 38.7 & \underline{46.9} & 44.2 & \textbf{58.3} \\
    \hline
    \end{tabular}}
    \label{tab:chestx8_zeroshot}
\end{table*}

\subsection{Optimization Pipeline}
\label{subsec:optimization}

To perform disease-oriented object detection, we integrate contrastive learning for region-text alignment with standard detection objectives for bounding box regression and classification. This unified optimization strategy enables the model to simultaneously learn discriminative visual representations aligned with disease semantics and accurate spatial localization of pathological abnormalities.

\subsubsection{Contrastive Region-Text Alignment}
Specifically, each spatial token in the enhanced image embeddings $X_{final}$ serves as a candidate region representation. The disease category embeddings from the disease beacon encoder serve as fixed semantic anchors in the shared embedding space. Each spatial token is projected into an object embedding $e_k$ and a bounding box prediction $\hat{b}_k$. We compute similarity scores $s_{k,j}$ between $e_k$ and each disease category embedding $w_j$ using cosine similarity. Our contrastive loss encourages alignment between visual regions and their corresponding disease categories:
\begin{equation}
\mathcal{L}_{\text{contrast}} = -\frac{1}{C} \sum_{k=1}^C \log \frac{\exp(s_{k,y_k})}{\sum_{j=1}^{N_{\text{cls}}} \exp(s_{k,j})},
\end{equation}
where $y_k$ denotes the ground-truth disease label associated with spatial location $k$, $C$ represents the number of positive samples, and $N_{\text{cls}}$ is the total number of disease classes. This contrastive objective ensures that visual features corresponding to specific diseases are pulled closer to their respective disease embeddings while being pushed away from irrelevant categories.

\subsubsection{Detection Head and Loss Function}
The predicted bounding boxes $\hat{b}_k$ and their associated class scores $\{s_{k,j}\}$ are processed by the standard detection head for final prediction. The detection loss combines IoU-based regression error and distribution-based distance \cite{li2022generalized}:
\begin{equation}
\mathcal{L}_{\text{bbox}} = \frac{1}{\sum_i s_i} \sum_{i} \left[(1 - \text{CIoU}(\hat{b}_i, b_i)) + \text{DFL}(p_i^{\text{dist}}, d_i^{\text{target}})\right] \cdot s_i,
\end{equation}
where $s_i$ denotes the predicted objectness score, $b_i$ represents the ground-truth bounding box, $\text{CIoU}$ denotes complete intersection over union for spatial alignment, and $\text{DFL}$ represents the distribution focal loss applied to discrete bounding box offsets for fine-grained localization. The total loss function combines the alignment objective and detection objective:
\begin{equation}
\mathcal{L}_{\text{total}} = \mathcal{L}_{\text{contrast}} + \mathcal{L}_{\text{bbox}}.
\end{equation}
This unified formulation enables the model to simultaneously align spatial features with disease semantics through contrastive learning and optimize detection performance through regression and classification objectives, achieving both semantic understanding and precise localization for multi-label lesion detection in chest X-rays.

\begin{table*}[!t]
\caption{Ablation study of the prompt generator; SCP: semantic context prompt, DBP: disease beacon prompt. }
\setlength\tabcolsep{16pt}
\resizebox{\textwidth}{!}{
\begin{tabular}{c|cccccccccc}
    \hline
    SCP & DBP & Precision & Recall & \textbf{$\rm AP_{\rm 40}$} & \textbf{$\rm AP_{\rm 50}$} & \textbf{$\rm AP_{\rm 60}$} & \textbf{$\rm AP_{\rm 70}$} & \textbf{$\rm AP_{\rm 80}$} & \textbf{$\rm AP_{\rm 90}$} & \textbf{$\rm mAP_{\rm 40:95}$} \\
    \hline
    & & 45.7 & 41.2 & 40.3 & 35.3 & 28.0 & 15.4 & \textbf{6.6} & \textbf{0.8} & 19.4 \\
    \checkmark & & \underline{46.2} & \underline{42.4} & \underline{41.2} & \underline{36.7} & \underline{28.2} & \underline{16.9} & 6.2 & 0.6 & \underline{20.2} \\
    & \checkmark & 45.9 & 41.8 & 39.2 & 36.1 & 28.1 & 15.9 & \underline{6.4} & 0.6 & 19.9 \\
    \checkmark & \checkmark & \textbf{46.4} & \textbf{43.8} & \textbf{42.4} & \textbf{38.2} & \textbf{29.5} & \textbf{18.7} & \textbf{6.6} & \underline{0.7} & \textbf{21.0} \\
    \hline
\end{tabular}
}
\label{tab:dual-text prompt}
\end{table*}

\begin{table*}[!t]
\caption{Performance comparison with different text encoders.}
\setlength\tabcolsep{16pt}
\resizebox{\textwidth}{!}{
\begin{tabular}{l| c c c c c c c c c c}
    \hline
    Text Encoders &Precision & Recall & \textbf{$\rm AP_{\rm 40}$} & \textbf{$\rm AP_{\rm 50}$} & \textbf{$\rm AP_{\rm 60}$} & \textbf{$\rm AP_{\rm 70}$} & \textbf{$\rm AP_{\rm 80}$} & \textbf{$\rm AP_{\rm 90}$} & \textbf{$\rm mAP_{\rm 40:95}$} \\
    \hline
    Bert-layer \cite{devlin2018bert} & 41.8 & 41.8 & 39.7 & 34.9 & 26.3 & \underline{17.6} & \underline{7.5} & 0.3 & 19.3 \\
    BioBert \cite{lee2020biobert}  & 38.8 & 41.5 & 38.4 & 33.7 & 27.3 & 16.5 & 6.4 & 0.4 & 18.7 \\
    BioClinicalBert \cite{alsentzer2019publicly} & \underline{43.6} & \underline{43.1} & \underline{40.3} & \underline{36.1} & \underline{27.5} & 17.5 & \textbf{7.7} & \underline{0.5} & \underline{19.9} \\
    Bert-base \cite{devlin2018bert} & \textbf{46.4} & \textbf{43.8} & \textbf{42.4} & \textbf{38.2} & \textbf{29.5} & \textbf{18.7} & 6.6 & \textbf{0.7} & \textbf{21.0} \\
    \hline
\end{tabular}
}
\label{tab:encoder_comparison}
\end{table*}

\begin{table*}[!t]
\caption{Analysis of depth in bidirectional feature enhancer.}
\setlength\tabcolsep{16pt}
\resizebox{\textwidth}{!}{
\begin{tabular}{c|cccccccccc}
    \hline
    Depth & Precision & Recall & \textbf{$\rm AP_{\rm 40}$} & \textbf{$\rm AP_{\rm 50}$} & \textbf{$\rm AP_{\rm 60}$} & \textbf{$\rm AP_{\rm 70}$} & \textbf{$\rm AP_{\rm 80}$} & \textbf{$\rm AP_{\rm 90}$} & \textbf{$\rm mAP_{\rm 40:95}$} \\
    \hline
    1 & \underline{44.0} & 39.6 & 38.3 & 34.0 & \underline{27.4} & \underline{18.6} & \textbf{8.2} & \underline{0.3} & \underline{19.3} \\
    2 & \textbf{46.4} & \underline{43.8} & \textbf{42.4} & \textbf{38.2} & \textbf{29.5} & \textbf{18.7} & \underline{6.6} & \textbf{0.7} & \textbf{21.0} \\
    3 & 40.2 & \textbf{44.7} & \underline{39.2} & \underline{35.0} & 26.2 & 16.3 & 6.2 & \underline{0.3} & 19.0 \\
    \hline
\end{tabular}
}
\label{tab:ablation}
\end{table*}

\section{Experiments}

\subsection{Datasets}
To comprehensively validate the effectiveness and generalization capability of the proposed SP-Det framework, we conduct extensive experiments on two widely-used chest X-ray datasets with diverse thoracic disease categories and varying annotation scales. The datasets are described as follows:

\subsubsection{VinDr-CXR} The VinDr-CXR dataset \cite{nguyen2022vindr}, derived from the VinBigData Chest X-ray Abnormalities Detection Kaggle challenge, comprises 4,394 fully-annotated posteroanterior chest X-ray scans with 14 distinct thoracic abnormalities, including pneumothorax, pleural effusion, nodule/mass, cardiomegaly, atelectasis, consolidation, interstitial lung disease (ILD), infiltration, lung opacity, pleural thickening, calcification, aortic enlargement, pulmonary fibrosis, and other lesions. These annotations cover both common and rare pathological conditions, where each scan is meticulously annotated by at least three board-certified radiologists with extensive clinical experience, ensuring high-quality and reliable ground truth labels. The dataset represents a diverse spectrum of patient demographics and disease manifestations, making it highly representative of real-world clinical scenarios. The VinDr-CXR dataset serves as our primary benchmark for comprehensive lesion detection evaluation.

\subsubsection{ChestX-ray8} 
The ChestX-ray8 dataset \cite{wang2017chestx}, developed by the National Institutes of Health (NIH) Clinical Center, is a large-scale benchmark for automatic thoracic disease detection and diagnosis. For our evaluation purposes, we utilize a subset containing 880 frontal-view CXR images from unique patients, annotated with 8 different thoracic lesions, including atelectasis, cardiomegaly, effusion, infiltration, mass, nodule, pneumonia, and pneumothorax. Since lesion-level bounding box annotations are labor-intensive and expensive in clinical applications, this dataset with limited annotations serves as an important testbed to demonstrate the domain adaptability and robustness of our framework under different annotation scales and distribution shifts. We employ this dataset to evaluate the generalization capability of SP-Det when trained on VinDr-CXR and tested on ChestX-ray8, simulating realistic cross-dataset transfer scenarios commonly encountered in clinical deployment.

\subsection{Evaluation Metrics}
To comprehensively assess detection performance, we adopt multiple standard evaluation metrics widely used in object detection literature. Specifically, we report (1) Precision and Recall to measure the trade-off between false positives and false negatives, (2) Average Precision (AP) at various Intersection-over-Union (IoU) thresholds, including $\text{AP}_{40}$, $\text{AP}_{50}$, $\text{AP}_{60}$, $\text{AP}_{70}$, $\text{AP}_{80}$, and $\text{AP}_{90}$, to evaluate localization accuracy at different strictness levels, and (3) mean Average Precision (mAP) computed over IoU thresholds from 0.40 to 0.95 with a step size of 0.05, following the COCO evaluation protocol, which provides a comprehensive measure of detection performance across varying localization requirements. For disease-specific analysis, we report $\text{AP}_{40}$ for each individual thoracic abnormality category to provide detailed insights into the model's performance on different pathological conditions.

\subsection{Experimental Setup} We conduct our experiment on the VinDr-CXR \cite{nguyen2022vindr}, which consists of 15,000 postero-anterior (PA) chest X-ray (CXR) scans in DICOM format, annotated with 14 types of abnormalities. The dataset is split into training, validation, and testing sets with a ratio of 8:1:1. Our experiments are performed on two NVIDIA A5000 GPUs with PyTorch. The model is optimized using the AdamW optimizer with an initial learning rate of $1.5\times10^{-4}$ and a batch size of 32. SP-Det was initialized with pre-trained weights transferred from YOLO-World, while the baseline models were fine-tuned from their respective pre-trained weights. The model is trained for a maximum of 100 epochs. Early stopping is employed to prevent overfitting, where training halts if the validation loss does not improve for 20 consecutive epochs. We adopt pre-trained CLIP ViT-B/32 \cite{radford2021learning} and BERT-base \cite{devlin2018bert} as text encoders. Since our method is built upon a YOLOv8s backbone, all compared models are of the \textit{small (S)} size to ensure fairness. Training is conducted using the Ultralytics\footnote{https://github.com/ultralytics/ultralytics} framework across all compared methods. The used CXR text generation model is CheXagent \cite{chen2024chexagent}, which does not include VinDr-CXR as the training set. We reported the mean AP (mAP) from $\rm AP_{40}$ to $\rm AP_{95}$ with an interval of 0.05 in our experiments. These settings are based on the existing challenge\footnote{https://www.kaggle.com/competitions/vinbigdata-chest-xray-abnormalities-detection} and studies \cite{gabruseva2020deep, luo2021oxnet}.

\subsection{Comparison with State-of-the-Art Methods}
To evaluate the performance of SP-Det, we compare it with state-of-the-art models on both overall detection metrics and category-level performance. As illustrated in Table~\ref{tab:model_comparison}, which reports results on the VinDr-CXR test set\cite{nguyen2022vindr}, SP-Det outperforms baselines across most metrics, notably improving $\rm AP_{\rm 40}$ and $\rm AP_{\rm 50}$ by more than 2\%, and achieving the highest recall (43.8\%) and $\rm mAP_{\rm 40:95}$ (21.0\%). Although YOLO-World slightly leads in precision and higher IoU thresholds, SP-Det remains highly competitive, consistently achieving the second-best performance in these metrics, which further demonstrates the effectiveness of integrating contrastive learning and textual information. Furthermore, Table~\ref{tab:disease_comparison} provides a fine-grained lesion-level comparison, revealing SP-Det's strong generalisation across 14 abnormality types. It ranks first in 8 categories, and achieves particularly high accuracy on Aortic enlargement, Cardiomegaly, and Pneumothorax (all above 85\%). These results confirm SP-Det’s robust generalisability and highlight the value of incorporating medical report texts for improved lesion detection. To further illustrate the effectiveness of SP-Det, Fig. \ref{fig:visualisation}  presents the qualitative visualization results on the VinDr-CXR test set. These examples highlight the model’s ability to accurately detect and localise multiple co-existing lesions within a single chest X-ray. SP-Det demonstrates strong performance in identifying a variety of abnormalities, even in complex cases with multi-disease occurrences. Notably, Aortic enlargement and Cardiomegaly are consistently detected across most samples, aligning well with the quantitative results reported in Table~\ref{tab:disease_comparison}. While minor deviations between the predicted bounding boxes and the ground truth are observed in a few cases, such as those in the second row, the model still successfully identifies the relevant lesion regions. 

\subsection{Generalization Analysis}
To validate the cross-domain generalization capability of our SP-Det, we evaluate all competing methods trained on VinDr-CXR dataset in a zero-shot manner on the ChestX-ray8 dataset \cite{wang2017chestx} for lesion detection. As shown in Table \ref{tab:chestx8_zeroshot}, our SP-Det demonstrates superior generalization performance across most disease categories, achieving an average AP of 33.5\% without any fine-tuning on the target domain. Notably, SP-Det achieves the best performance in 6 out of 8 disease categories, with particularly significant improvements in Pneumothorax (58.3\% vs. 46.9\%), Cardiomegaly (63.9\% vs. 60.1\%), and Atelectasis (12.8\% vs. 10.3\%). The substantial gain in Pneumothorax detection (+11.4\%) is especially noteworthy, as this condition often requires precise localization of subtle air-filled spaces, demonstrating the effectiveness of our semantic context prompts in capturing fine-grained pathological features. While our method shows competitive but not optimal performance in Infiltration and Mass detection, the overall results confirm the effectiveness of our self-prompting mechanism in cross-domain scenarios. The consistent performance advantages demonstrate that SP-Det's dual-text prompt generation and bidirectional feature enhancement provide robust semantic representations that generalize well to unseen chest X-ray domains, highlighting its potential for practical deployment.

\subsection{Ablation Study}
To further investigate the effectiveness of key components in SP-Det, we conduct a series of ablation studies, including the design of expert-free dual-text prompt generator, the choice of text encoder, and the number of feature enhancement layers. As shown in Table ~\ref{tab:dual-text prompt}, incorporating both SCP and DBP  leads to the best performance except for $\rm AP_{\rm 90}$. Using only SCP yields the second-best results, while using only DBP still outperforms the baseline with neither prompt. These results confirm that both components contribute positively, and their combination provides complementary benefits for improved lesion detection. In terms of text encoder selection (Table~\ref{tab:encoder_comparison}), BERT-base achieves the best overall performance across most evaluation metrics, with the exception of $\rm AP_{\rm 80}$. While Bio\_ClinicalBERT attains the highest $\rm AP_{\rm 80}$ (7.7\%), it performs worse than BERT-base on key metrics such as $\rm AP_{\rm 40}$ and $\rm AP_{\rm 50}$. Other variants, including BERT-layer and BioBERT, show relatively lower performance, with the lowest mAP scores among all evaluated models. These findings suggest that the general-purpose BERT-base encoder is more effective than domain-specific alternatives for integrating textual information in this lesion detection task. Additionally, Table~\ref{tab:ablation}  compares SP-Det models with different numbers of feature enhancement layers. Using two layers yields the best overall performance, ranking first on most evaluation metrics and second on a few others. In particular, compared to using one or three layers, the two-layer configuration improves precision by at least 5\%, and $\rm mAP_{\rm 40:95}$ by over 8\%. These results indicate the existence of an optimal enhancement depth, beyond which additional stacking may introduce redundancy and hinder performance. Overall, these ablation results validate the effectiveness of our design choices and highlight the contribution of each component to SP-Det.

\subsection{Discussion}
While our framework demonstrates promising automated performance, its integration into existing clinical workflows requires further investigation. Our approach inherently relies on the quality of vision-language model outputs for prompt generation, and although our post-processing pipeline mitigates common generation issues, performance remains bounded by the underlying VLM's medical knowledge and capabilities. The disease beacon extraction process depends on natural language processing techniques for medical entity recognition, which may introduce domain-specific challenges. Additionally, the framework's scalability to different imaging modalities and disease categories warrants further investigation. Despite these considerations, SP-Det represents a significant step toward practical automated lesion detection by eliminating manual annotation while maintaining competitive performance.

\section{Conclusion}
In this work, we introduce SP-Det, a novel self-prompting detection framework designed to enhance performance in multi-label chest X-ray lesion detection without requiring labor-intensive manual annotations. The core innovations of our framework include an expert-free dual-text prompt generator that creates complementary textual guidance. Furthermore, our bidirectional feature enhancer synergistically integrates these textual representations with visual features, significantly improving feature representation capabilities. Extensive experiments validate the effectiveness of our approach, demonstrating a significant advancement over existing state-of-the-art methods in multi-label chest X-ray lesion detection.

\section*{Declaration of competing interest}
The authors declare that they have no known competing financial interests or personal relationships that could have
appeared to influence the work reported in this paper.

\section*{Acknowledgements}
This work is partially supported by the Yongjiang Technology Innovation Project (2022A-097-G), Zhejiang Department of Transportation General Research and Development Project (2024039), and National Natural Science Foundation of China grant (UNNC ID: B0166).

\bibliographystyle{unsrt}
\bibliography{cleanbib}
	
\end{document}